\documentclass[letterpaper]{article} 
\usepackage{aaai23}  
\usepackage{times}  
\usepackage{helvet}  
\usepackage{courier}  
\usepackage[hyphens]{url}  
\usepackage{graphicx} 
\usepackage{natbib}  
\usepackage{caption} 
\usepackage{wrapfig}
\usepackage{graphicx}
\usepackage{subfig}
\usepackage{multirow}
\usepackage{booktabs}
\usepackage{stfloats}
\usepackage{color}
\usepackage{xcolor}
\usepackage{amsthm}
\usepackage{amsmath}
\usepackage{amsfonts}
\usepackage{algorithm}
\usepackage{algorithmic}

\newcommand{\pink}[1]{\textcolor[RGB]{220,27,116}{#1}}
\newcommand{\anno}[1]{\textcolor[RGB]{65,128,128}{#1}}

\def\HiLi{\leavevmode\rlap{\hbox to \linewidth{\color{yellow!3}\leaders\hrule height 0.8\baselineskip depth .5ex\hfill}}}

\usepackage{newfloat}
\usepackage{listings}
\DeclareCaptionStyle{ruled}{labelfont=normalfont,labelsep=colon,strut=off} 
\floatstyle{ruled}
\newfloat{listing}{tb}{lst}{}
\floatname{listing}{Listing}
\title{Contrastive Identity-Aware Learning for Multi-Agent Value Decomposition}
\author {
    Shunyu Liu\equalcontrib, 
    Yihe Zhou\equalcontrib, 
    Jie Song\thanks{Corresponding author.}, 
    Tongya Zheng, 
    Kaixuan Chen,\\
    Tongtian Zhu, 
    Zunlei Feng, 
    Mingli Song
}
\affiliations {
    Zhejiang University\\
    liushunyu@zju.edu.cn, yihe\_zhou@zju.edu.cn, sjie@zju.edu.cn, tyzheng@zju.edu.cn, chenkx@zju.edu.cn, raiden@zju.edu.cn, zunleifeng@zju.edu.cn, brooksong@zju.edu.cn
}

\begin{document}

\maketitle

\begin{abstract}
    Value Decomposition~(VD) aims to deduce the contributions of agents for decentralized policies in the presence of only global rewards, and has recently emerged as a powerful credit assignment paradigm for tackling cooperative Multi-Agent Reinforcement Learning~(MARL) problems. One of the main challenges in VD is to promote diverse behaviors among agents, while existing methods directly encourage the diversity of learned agent networks with various strategies. However, we argue that these dedicated designs for agent networks are still limited by the indistinguishable VD network, leading to homogeneous agent behaviors and thus downgrading the cooperation capability. In this paper, we propose a novel Contrastive Identity-Aware learning~(CIA) method, explicitly boosting the credit-level distinguishability of the VD network to break the bottleneck of multi-agent diversity. Specifically, our approach leverages contrastive learning to maximize the mutual information between the temporal credits and identity representations of different agents, encouraging the full expressiveness of credit assignment and further the emergence of individualities. The algorithm implementation of the proposed CIA module is simple yet effective that can be readily incorporated into various VD architectures. Experiments on the SMAC benchmarks and across different VD backbones demonstrate that the proposed method yields results superior to the state-of-the-art counterparts. Our code is available at \url{https://github.com/liushunyu/CIA}.
\end{abstract}

\section{Introduction}

Cooperative Multi-agent Reinforcement Learning~(MARL) aims to jointly train multiple agents to achieve one common goal, and has witnessed its unprecedented success in various applications, such as video games~\cite{AlphaStar,FTW}, traffic light systems~\cite{wu2020multi}, and smart grid control~\cite{yan2020multi,xu2020multi,wang2021multi}. To alleviate the partial observation constraint and the scalability problem in cooperative MARL, the \emph{Centralized Training and Decentralized Execution}~(CTDE) framework has attracted increasing attention~\cite{MADDPG}, where agents are granted access to additional global information during centralized training and deliver actions only based on their local histories in a decentralized way. The decentralized policies with one shared network significantly reduce the number of learnable parameters and the exponentially growing joint-action space to the linear complexity. Nevertheless, the CTDE framework still faces a critical challenge of credit assignment, \emph{i.e.,} how to deduce the contributions of individual agents given only global rewards.

\begin{figure}[!t]
    \centering
    \includegraphics[width=0.47\textwidth]{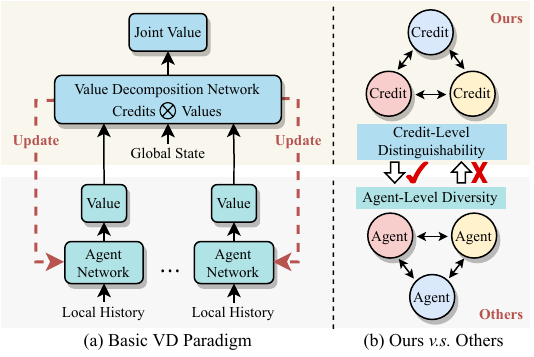}
    \caption{(a) Basic VD paradigm. $\bigotimes$ denotes the complex network operator based on different VD methods, where credits can be considered as the agent contributions. (b) Comparing the credit-level distinguishability mechanism with the agent-level diversity mechanism under the VD paradigm. Credit-level distinguishability can promote agent-level diversity, but not vice versa.}
    \label{fig:demo}
\end{figure}

In recent years, Value Decomposition~(VD) has emerged as a promising credit assignment paradigm, which endows agents with the ability to learn their own optimal policies by decomposing the joint value function according to their individual credits, as depicted in Figure~\ref{fig:demo}(a). The seminal work, VDN~\cite{VDN}, additively factorizes the team value function into individual agent-wise terms, imposing a strong constraint of credit equalization. QMIX~\cite{QMIX} identifies that the full factorization of VDN is not necessary and relaxes this additive constraint to a non-linear combination, which enables the weighted credits for per-agent values. Due to the superior performance of QMIX, there have been many recent efforts to improve its representation capability~\cite{WQMIX,QPLEX}.

However, despite the large representation capacity of existing VD methods, it is widely observed that agents often learn similar behaviors inevitably~\cite{MAVEN,hu2022policy}. 
Such similar behaviors easily lead to the local optimum of the cooperative policies, which may severely impede the efficient exploration and downgrade the final performance~\cite{terry2020revisiting}. Consider a football game where different agents observe similarly. If agents behave similarly for desired credits, all of them may gather to compete for a ball. However, they should keep distance and adopt diverse tactics. Several studies attribute this similarity between agents to the homogeneous policy networks with parameter sharing~\cite{EOI,CDS}. They propose to design auxiliary objectives in policy networks towards agent-level diversity, and even introduce agent-specific networks for each agent while sacrificing the advantage of complete parameter sharing. However, these works ignore the fact that policy networks are evaluated and improved via the VD network, where the diversity of agent behaviors often depends on the distinguishability of credit assignment, as depicted in Figure~\ref{fig:demo}(b). Thus, we argue that their pursuit of diverse policies is still limited when the VD network provides ambiguous credits to demonstrate the contributions of different agents.

In this paper, we investigate the multi-agent diversity from a novel perspective of credit assignment. To evaluate the distinguishability of existing VD methods, we design a random-shuffle scheme that eliminates the identity information of input agent values while preserving the original network architecture. Case studies empirically show that the VD network may assign ambiguous credits to the agents, indicating that promoting the diversity of agents will be limited by the identity insensitivity of the VD network. Therefore, we introduce a new contrastive identity-aware learning method, termed as CIA, to explicitly encourage credit-level distinguishability. The proposed method leverages gradient-based attribution to represent the credit of each agent, and further considers the long-term behaviors of agents by adopting the temporal credits from the overall trajectory.
To encourage the discriminative assignment of credits and further the emergence of diverse behaviors, we propose to maximize the mutual information between the temporal credits and learnable identity representations of different agents. Moreover, we customize a contrastive learning objective to derive a tractable lower bound for the mutual information since estimating and maximizing the mutual information of neural networks is often intractable. Our main contributions can be summarized as follows: 
\begin{itemize}
    \item We identify the ambiguous credit assignment problem in VD, a highly important ingredient for multi-agent diversity yet largely overlooked by existing literature. 
    \item We propose a novel contrastive identity-aware learning~(CIA) method to promote diverse behaviors via explicitly encouraging credit-level distinguishability. The proposed CIA module imposes no constraints over the network architecture, and serves as a plug-and-play module readily applicable to various VD methods.
    \item Experiments conducted on the StarCraft II micromanagement benchmark show that CIA yields significantly superior performance to state-of-the-art competitors.
\end{itemize}

\section{Related Work}

\paragraph*{Value Decomposition} aims to extract the individual utility from the global reward for credit assignment, which has become a well-established paradigm for tackling cooperative MARL problems~\cite{zohar2022locality,jeon2022maser,fu2022revisiting}. To realize efficient VD, it is critical to satisfy the Individual-Global-Max~(IGM) principle that the global optimal action should be consistent with the collection of individual optimal actions of agents~\cite{QTRAN}. 
Following the IGM principle, VDN~\cite{VDN} proposes to represent the joint value function as a sum of individual value functions, while QMIX~\cite{QMIX} extends this additive VD and imposes a monotonicity constraint, showing the state-of-the-art performance. To further alleviate the risk of suboptimal results, WQMIX~\cite{WQMIX} improves QMIX by a weighted projection that allows more emphasis to be placed on underestimated actions. 

However, VDN and QMIX still suffer from structural constraints which limit their representation capability for joint action-value function classes. Thus, QTRAN~\cite{QTRAN} constructs the soft regularization constraints to achieve more general decomposition than VDN and QMIX. QPLEX~\cite{QPLEX}, on the other hand, proposes a duplex dueling network architecture to enable the complete VD function class that satisfies the IGM principle. These VD methods have shown great potential in the field of credit assignment for challenging MARL tasks. Other works further adapt VD towards transfer learning~\cite{DBLP:conf/iclr/LongZ0FWW20,DyMA,yang2022deep,yang2022factorizing} and ad hoc teamwork~\cite{gu2021online,DBLP:conf/aaai/MackeMS21,DBLP:conf/icml/RahmanHCA21}. Despite the success of VD, existing methods ignore the distinguishability in credit assignment, leading to the homogeneous behaviors among multiple agents. GRE~\cite{GRE} tries to use credit entropy regularization to mitigate this problem, which only imposes single-step distinguishability on credits.

\paragraph*{Agent Diversity} has been widely studied in single-agent reinforcement learning problems, which provides an exploration bonus to encourage the diverse behaviors~\cite{eysenbach2018diversity,DBLP:conf/iclr/BurdaESK19}. In recent years, due to the promising results achieved by the existing single-agent methods, diversity has also emerged as a popular topic in MARL~\cite{lee2019learning,christianos2021scaling}. It encourages the difference between individual agents under the cooperative setting when pursuing diversity in the context of MARL. RODE~\cite{RODE} realizes a learnable role assignment for agents to achieve diversity. It learns action representations and clusters different actions into several restricted roles, while this role-based method is limited by the action space that cannot be decomposed. EOI~\cite{EOI} proposes to construct the intrinsic reward, a predicted probability of agent identity given its observation, to promote the emergence of individuality. However, the identity prediction mechanism of EOI is easy to overfit if the local observation contains the identity information. In MAVEN~\cite{MAVEN}, agents condition their behaviors on the shared latent variable controlled by a hierarchical policy, where the mutual information between the trajectories and latent variables is maximized to learn a diverse set of such behaviors. Similarly, EITI~\cite{DBLP:conf/iclr/0001WWZ20} leverages mutual information to capture the influence between the transition dynamics of different agents, while CDS~\cite{CDS} designs agent-specific networks for each agent and encourages diversity by optimizing the mutual information between the agent identities and trajectories. These works are all built based on the existing VD methods, especially QMIX, encouraging only the diversity of agent networks with various strategies. Thus, they still suffer from limited distinguishability when credit assignment in the VD network is ambiguous.

\section{Preliminary}

\paragraph*{Dec-POMDP.}
We consider a fully cooperative multi-agent setting under the \textit{Decentralized Partially Observable Markov Decision Process}~(\text{Dec-POMDP}), which is defined as a tuple $\langle \mathcal{A},\mathcal{S},\mathcal{U},P,r,\Omega,O,\gamma \rangle$, where $\mathcal{A} = \{a_k\}_{k=1}^K$ is the set of $K$ agents and $s\in\mathcal{S}$ is the global state of the environment. At each time step $t$, each agent $a_k \in \mathcal{A}$ receives an individual partial observation $o^k_t \in \Omega$ drawn from the observation function $O(s_t, a_k)$. Then each agent chooses a corresponding action $u^k_t \in \mathcal{U}$, forming a joint action $\boldsymbol{u}_t \in \mathcal{U}^K$. This causes a transition to the next state $s_{t+1}$ according to the state transition function $P(s_{t+1} | s_t, \boldsymbol{u}_t):\mathcal{S}\times\mathcal{U}^K\times\mathcal{S} \to [0,1]$. All agents share the same reward function $r(s_t, \boldsymbol{u}_t):\mathcal{S}\times\mathcal{U}^K\to \mathbb{R}$ and $\gamma \in [0, 1)$ is the discount factor. Each agent $a_k$ has an action-observation history $\tau^k \in \mathcal{T} \equiv (\Omega \times \mathcal{U})^*$ and learns its individual policy $\pi^k(u^k|\tau^k):\mathcal{T}\times\mathcal{U}\to[0,1]$ to jointly maximize the discounted return $R_t = \sum_{i=0}^{\infty}{\gamma^i r_{t+i}}$. The joint action-observation history is define as $\boldsymbol{\tau}\in \mathcal{T}^K$. The joint policy $\boldsymbol{\pi}$ induces a joint action-value function $Q^{tot}_t(s_t,\boldsymbol{u}_t)=\mathbb{E}_{s_{t+1:\infty},\boldsymbol{u}_{t+1:\infty}}{\left[R_t \mid s_t,\boldsymbol{u}_t, {\boldsymbol{\pi}}\right]}$ that represents the expected discounted return under the given policy.

\paragraph*{The CTDE Framework} has attracted substantial attention in cooperative MARL to achieve effective policy learning~\cite{yu2021surprising,DBLP:conf/kdd/LuoL0KLSW22}, where agents must learn the decentralized policies which based on only local observation at execution time, but they are granted access to the global information during centralized training. One of the promising ways to exploit the CTDE framework is VD, which allows agents to learn their individual utility functions by optimizing the joint action-value function for credit assignment. To realize VD, a mixing network with parameters $\theta_{\upsilon}$ is adopted as an approximator to estimate the joint action-value function $Q^{tot}$. 
The mixing network is introduced to merge all individual action values into a joint one $Q^{tot} = f(\boldsymbol{q};\theta_{\upsilon })$, where $\boldsymbol{q} = [Q^k]_{k=1}^K \in \mathbb{R}^K$ and $Q^k$ with shared parameters $\theta_{\pi}$ is the utility network of each agent $a_k$. The learnable parameter $\theta=\{\theta_{\pi},\theta_{\upsilon }\}$ can be updated by minimizing the following Temporal-Difference~(TD) loss:
\begin{align}
    \mathcal{L}_{TD}(\theta) =  \mathbb{E}_{\mathcal{D}} \left[\left(y^{tot} - Q^{tot}\right)^2\right].
\end{align}
where $\mathbb{E}[\cdot]$ denotes the expectation function, $\mathcal{D}$ is the replay buffer of the transitions, $y^{tot}=r+\gamma \hat{Q}^{tot}$ is the one-step traget and $\hat{Q}^{tot}$ is the target network~\cite{DQN15}.

\section{Method\label{sec:method}}

In what follows, we provide case studies to evaluate the credit indistinguishability of existing VD methods. Then we further detail the proposed contrastive identity-aware learning (CIA) module and finally summarize the complete optimization algorithm.

\begin{figure}[!b]
    \centering
    \subfloat[2c\_vs\_64zg (Hard)]{\includegraphics[width=0.25\textwidth]{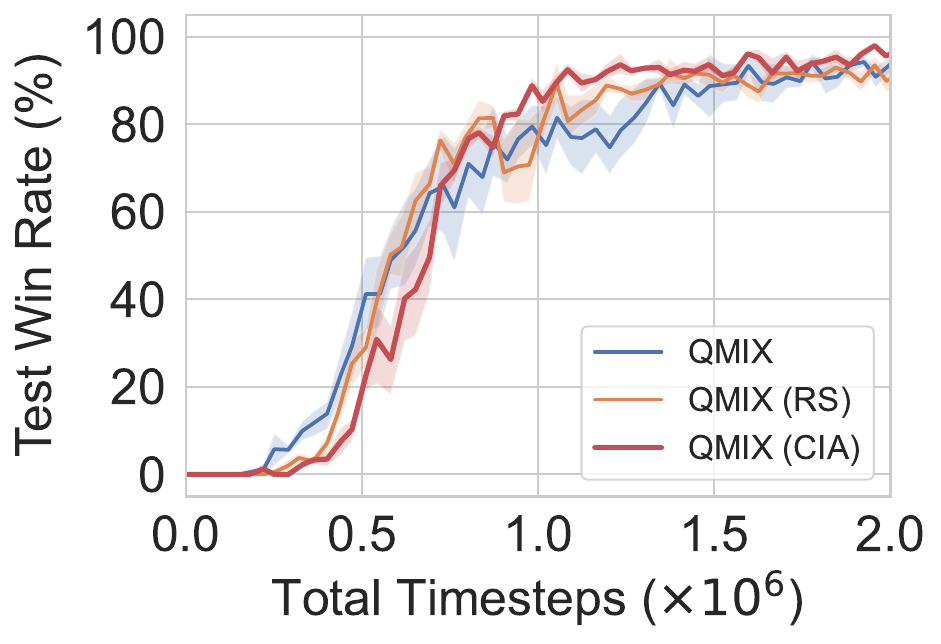}}\,
    \subfloat[KL Matrix (2c\_vs\_64zg)]{\includegraphics[width=0.215\textwidth]{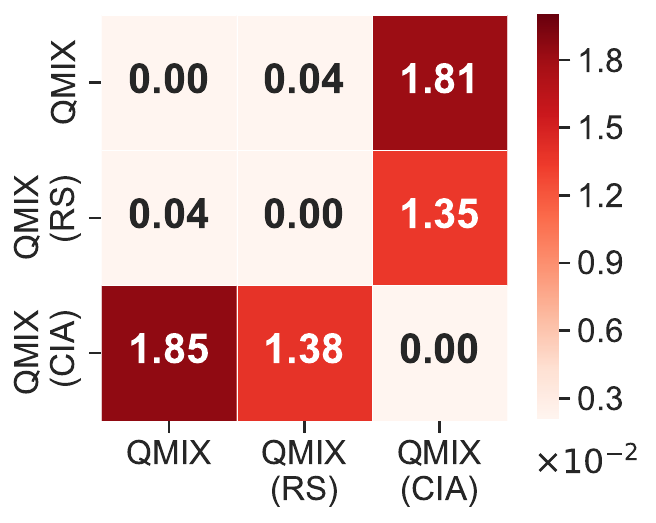}}\\
    \subfloat[3s5z\_vs\_3s6z (Super Hard)]{\includegraphics[width=0.25\textwidth]{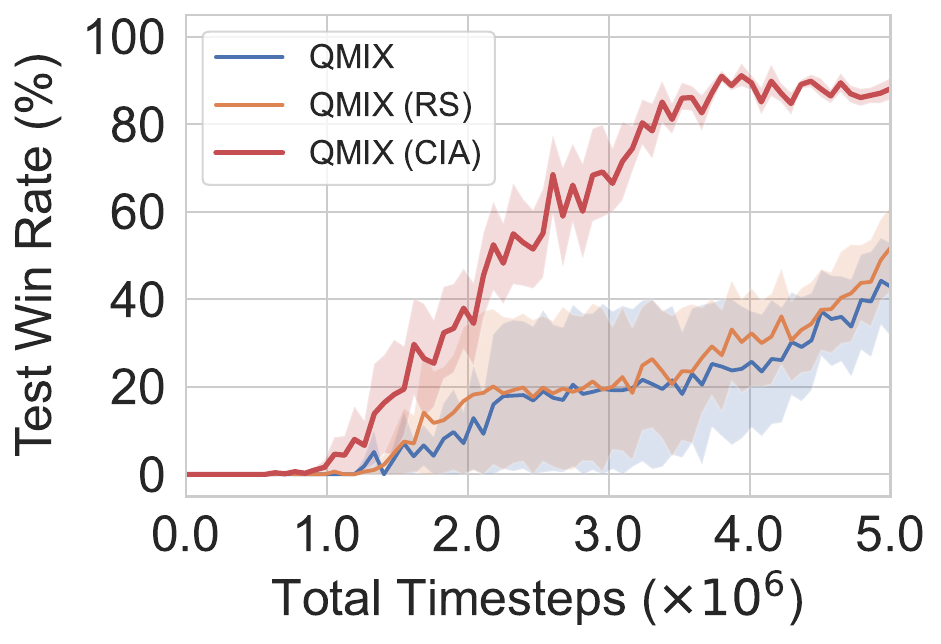}}\,
    \subfloat[KL Matrix (3s5z\_vs\_3s6z)]{\includegraphics[width=0.215\textwidth]{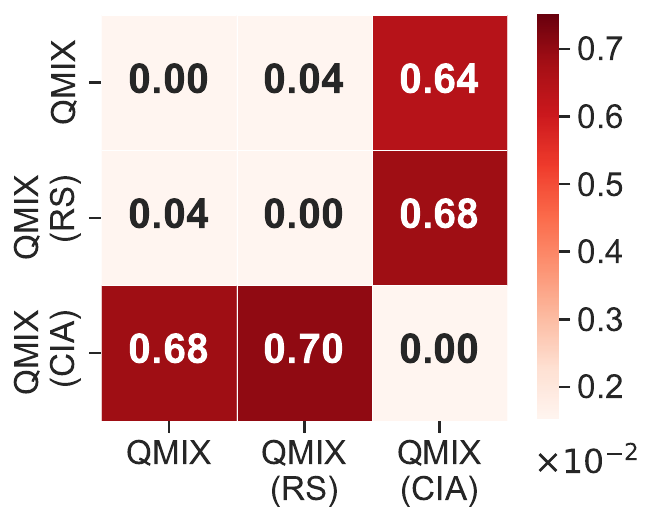}}
    \caption{(Left) Learning curves of QMIX and its variants on two SMAC scenarios. (Right) KL-divergence distance matrices of the credit distributions of different methods.}
    \label{fig:case}
  \end{figure}

\subsection{Credit Indistinguishability Analysis}

Firstly, we define that a credit assignment is ambiguous if the learnable credits are invariant to the agent identities. To investigate the ambiguous credit assignment problem in VD, we design a training scheme that randomly shuffles the order of input values to the mixing network at every training epoch. Concretely, we denote $\Phi$ as the set of all permutation matrices $\boldsymbol{P}\in\mathbb{R}^{K\times K}$. At each training epoch, we randomly sample a permutation matrix $\boldsymbol{P} \in \Phi$ and obtain the ambiguous joint-action value $Q^{tot} = f(\boldsymbol{P}\boldsymbol{q})$ instead of the original one to calculate the TD loss and update the network parameter. This random-shuffle scheme eliminates the identity information of input agent values without changing the original network architecture, providing a comparable baseline with ambiguous credit assignment. Furthermore, we propose to use KL-divergence distance to measure the similarity of credit distributions\footnote{At each time step $t$, the credit distribution $\boldsymbol{d}_t$ is defined as the normalized gradient-based credits (mentioned in Eq.~\ref{eq:credit}) over all agents: $\boldsymbol{d}_t = \operatorname{softmax}\left( \left[x^1_t,x^2_t,\cdots,x^K_t\right] \right)$.} of different methods. We calculate the average KL-divergence distance by the trajectories sampled by the trained networks of all compared methods separately. Please refer to Appendix B for more details about the experiment settings.

Case studies are conducted on two challenging tasks provided by SMAC~\cite{SMAC}. We take QMIX~\cite{QMIX} as an example to study due to its state-of-the-art performance, and introduce its variant under the random-shuffle training scheme, namely QMIX~(RS). Moreover, we also integrate the proposed CIA module with QMIX, namely QMIX~(CIA), to demonstrate the importance of credit-level distinguishability. Experimental results of different methods are shown in Figure~\ref{fig:case}. Interestingly, even with the ambiguous credit assignment, the random-shuffle variant can still achieve the performance on par with those obtained by original QMIX. The KL-divergence distance also shows that the credit distribution of QMIX is similar as its random-shuffle variant, indicating that QMIX are insensitive to the identity of agents. Thus, its learned credits may be ambiguous, which limits the diverse behaviors of agents and damages the final performance.

\begin{figure*}[!t]
    \centering
    \includegraphics[width=1.0\textwidth]{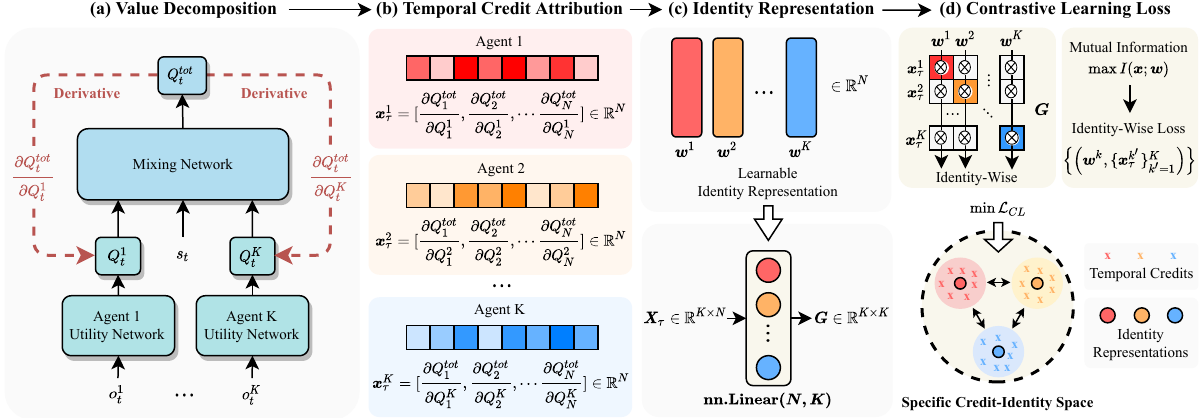}
    \caption{An illustrative diagram of the proposed contrastive identity-aware learning (CIA) method.}
    \label{fig:framework}
  \end{figure*}

\subsection{Contrastive Identity-Aware Learning}

To encourage multi-agent diversity, we introduce a novel contrastive learning method to realize the identity-aware distinguishability in credit assignment, as shown in Figure~\ref{fig:framework}.

\paragraph*{Temporal Credit Attribution.} Before promoting the distinguishability of credits, one key issue to be considered is: how to define a universal credit to represent the contributions of individual agents in different VD methods. Towards addressing this problem, we consider a gradient-based attribution mechanism to extract the credit assignment information in the mixing network. In general, the gradient-based attribution is calculated taking the partial derivatives of the output with respect to the input, indicating the relative importance of different inputs to a specific output~\cite{DBLP:conf/iclr/AnconaCO018}. Thus, we employ this mechanism to attribute the credits as the partial derivatives between the joint-action value and individual values, as shown in Figure~\ref{fig:framework}(a). Specially, the credit of the agent $a_k$ at each time step $t$ is formulated as: 
\begin{align}
    x^k_t = \frac{\partial Q^{tot}_t}{\partial Q^k_t} \in \mathbb{R}.\label{eq:credit}
\end{align}
This credit attribution mechanism enables us to determine the contribution of each agent value, also known as the value sensitivity. That is to say, a larger attribution magnitude means that the corresponding agent has a significant impact on the final result. Furthermore, this credit attribution mechanism can be readily applied to different VD methods, regardless of the heterogeneous mixing network architectures.

In sequential decision-making problem, it is insufficient to deduce the contribution of an agent only by the behavior of a single step. The behavior of an agent is usually affected by its final goal. As shown in Figure~\ref{fig:framework}(b), we therefore adopt the temporal credit attribution of each agent $a_k$ from the overall sampled trajectory $\tau$ with $t \in \{1,2,\cdots,N\}$ as:
\begin{align}
    \boldsymbol{x}^k_{\tau} = \left[x^k_1,x^k_2,\cdots,x^k_N\right]\in\mathbb{R}^N,
\end{align}
where $N$ is the maximum length of the sampled trajectories. If any trajectories are truncated early due to the game rules, zero padding will be applied to align the lengths of all trajectories. We denote $\boldsymbol{X}_{\tau} \in \mathbb{R}^{K\times N}$ as the temporal credits of $K$ agents in the trajectory $\tau$. 
This temporal credit attribution considers the long-term behaviors of agents, leading to a more stable measure for the agent contributions.

\paragraph*{Contrastive Identity-Aware Distinguishability.} 

After defining the credits in a reasonable way, we further attempt to promote the distinguishability of the credits, avoiding capacity degradation of the VD model due to the ambiguous credit assignment problem. However, obtaining the solution that all the credits differ from each other is not trivial. It is hard to directly constrain the distance between different learnable credits during the optimization process. Thus, we approximate the solution by introducing latent identity representations for each agent as intermediate variables, and condition the individual temporal credits on these representations, realizing identity-aware distinguishability.

Specially, each agent $a_k$ is assigned only one learnable random-initialized identity representation $\boldsymbol{w}^k \in \mathbb{R}^N$ with the same dimension of $N$ as the temporal credits during the entire training process. We denote $\boldsymbol{W} \in \mathbb{R}^{K\times N}$ as the identity representations of $K$ agents. On the other hand, it is notable that the temporal credits $\boldsymbol{x}^k_{\tau}$ of the agent are various in each sampled trajectory $\tau$. We propose to maximize the mutual information $I(\boldsymbol{x}; \boldsymbol{w})$ between the temporal credits and identity representations of different agents:
\begin{align}
    I(\boldsymbol{x}; \boldsymbol{w}) = \mathbb{E}_{\boldsymbol{x}, \boldsymbol{w}}\left[ \log \frac{p(\boldsymbol{x}\mid\boldsymbol{w})}{p(\boldsymbol{x})} \right].
\end{align}

However, directly optimizing this objective is quite difficult since estimating mutual information is often intractable. Inspired by contrastive learning~\cite{oord2018representation}, we introduce a contrastive learning loss, the InfoNCE loss, to provide a traceable lower bound of the mutual information objective as follows:
\begin{align}
    I(\boldsymbol{x}; \boldsymbol{w}) \geq \log(K) - \mathcal{L}_{CL},
\end{align}
where $\mathcal{L}_{CL}$ is the InfoNCE loss and $K$ is the number of agents. 
Contrastive learning can be considered as learning a differentiable dictionary look-up task, which contrasts semantically similar and dissimilar query-key pairs.
To match individual temporal credits with their corresponding identity representations, we define identity representations as queries and temporal credits as keys. For example, given a query $\boldsymbol{w}^{k}$ and keys $\mathcal{X} = \{\boldsymbol{x}^{k'}_{\tau}\}_{k'=1}^K$, the goal of contrastive learning is to ensure that $\boldsymbol{w}^{k}$ is close with $\boldsymbol{x}^{k}_{\tau}$ while being irrelevant to other keys in $\mathcal{X} \backslash \{\boldsymbol{x}^{k}_{\tau}\}$. Specially, the identity-wise contrastive learning loss is given as follows:
\begin{align}
    \mathcal{L}_{CL} = \mathop{\mathbb{E}}\limits_{ \left(\boldsymbol{w}^{k},\{\boldsymbol{x}^{k'}_{\tau}\}_{k'=1}^K\right) \sim \mathcal{D}}  \left[ - \log \frac{\exp(g(\boldsymbol{x}^{k}_{\tau},\boldsymbol{w}^{k}))}{\sum_{k'=1}^{K} \exp(g(\boldsymbol{x}^{k'}_{\tau},\boldsymbol{w}^{k}))} \right],
\end{align}
where $g(\boldsymbol{x}_\tau,\boldsymbol{w}) = \boldsymbol{x}^T_\tau \boldsymbol{w}\in\mathbb{R}$ is the function that compute the non-negative similarity score between the identity representation $\boldsymbol{w}$ and the temporal credit $\boldsymbol{x}_\tau$. Here we adopt a simple dot-product similarity, but it can be easily replaced by other methods. The similarity matrix is denoted~as $\boldsymbol{G} = \boldsymbol{X}_\tau \boldsymbol{W}^T\in\mathbb{R}^{K\times K}$. As shown in Figure~\ref{fig:framework}(d), this identity-wise contrastive learning loss constrains the learned identity representations to uniformly distribute on a specific credit-identity hypersphere without divergence, where the temporal credits distribute around their corresponding identity representation~\cite{wang2020understanding}. With the contrastive learning loss, CIA directly encourages the identity-aware distinguishability among agent credits, providing a more discriminative credit assignment for multi-agent diversity.

Intuitively, we can explain the intrinsic mechanism in CIA from the perspective of a classification problem, where each input credits predict their corresponding identity labels. The identity representations are the weights of the classifier. For classification, there are two ways to obtain a minimal loss: use some easier inputs or adopt a more complex classifier. Here the identity representations only form a simple one-layer linear classifier, which imposes a strong constraint that the input credits must be linearly separable for minimal loss. Thus, the CIA module successfully endows the VD network the ability to promote credit distinguishability.

\subsection{Optimization Algorithm}

To sum up, training our framework based on the CIA module contains two main parts. The first one is the original TD loss $\mathcal{L}_{TD}$, which enables each agent to learn its individual policy by optimizing the joint-action value of the mixing network. The second one is the proposed contrastive learning loss $\mathcal{L}_{CL}$ to facilitate the credit-level distinguishability. Thus, given these two corresponding loss items, the total loss of our framework is formulated as follows:
\begin{align}
    \mathcal{L}_{all} = \mathcal{L}_{TD} + \alpha \mathcal{L}_{CL},
\end{align}
where $\alpha$ is the coefficient for trading off loss terms. The overall framework is trained in an end-to-end centralized manner. It is notable that CIA does not break any constraints of original VD paradigm and still follows IGM principle. To make the proposed CIA clearer to readers, we provide the pseudocode in Appendix A. We only need to introduce an additional linear layer for learnable identity representations as shown in Figure~\ref{fig:framework}(c). The algorithm implementation of the CIA module is simple yet effective that can be seamlessly integrated with various VD architectures.

\section{Experiments}

To demonstrate the effectiveness of the proposed CIA method, we conduct experiments on the didactic game and the StarCraft II micromanagement challenge.

\subsection{Experimental Settings}

Our methods are compared with various state-of-the-art methods, including (i) Value decomposition methods: QMIX~\cite{QMIX}, QPLEX~\cite{QPLEX}, QTRAN~\cite{QTRAN}, OWQMIX and CWQMIX~\cite{WQMIX}. (ii) Diversity-based methods: MAVEN~\cite{MAVEN}, EOI~\cite{EOI}, GRE~\cite{GRE}, SCDS that is a variant of CDS~\cite{CDS} with complete shared agent network to ensure comparability. Our CIA implementation uses QMIX and QPLEX as the integrated backbones to evaluate its performance, namely QMIX~(CIA) and QPLEX~(CIA). These two methods are chosen for their robust performance in different multi-agent tasks, while CIA can also be readily applied to other methods.
We adopt the Python MARL framework (PyMARL)~\cite{SMAC} to implement our method and all baselines. The detailed hyperparameters are given in Appendix B, where the common training parameters across different methods are consistent.

\begin{figure*}[!t]
    \centering
    \subfloat[Game Visualization]{\includegraphics[width=0.245\textwidth]{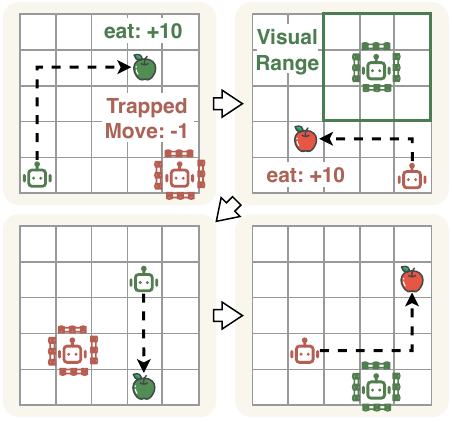}}\;
    \subfloat[Learning Curves]{\includegraphics[width=0.33\textwidth]{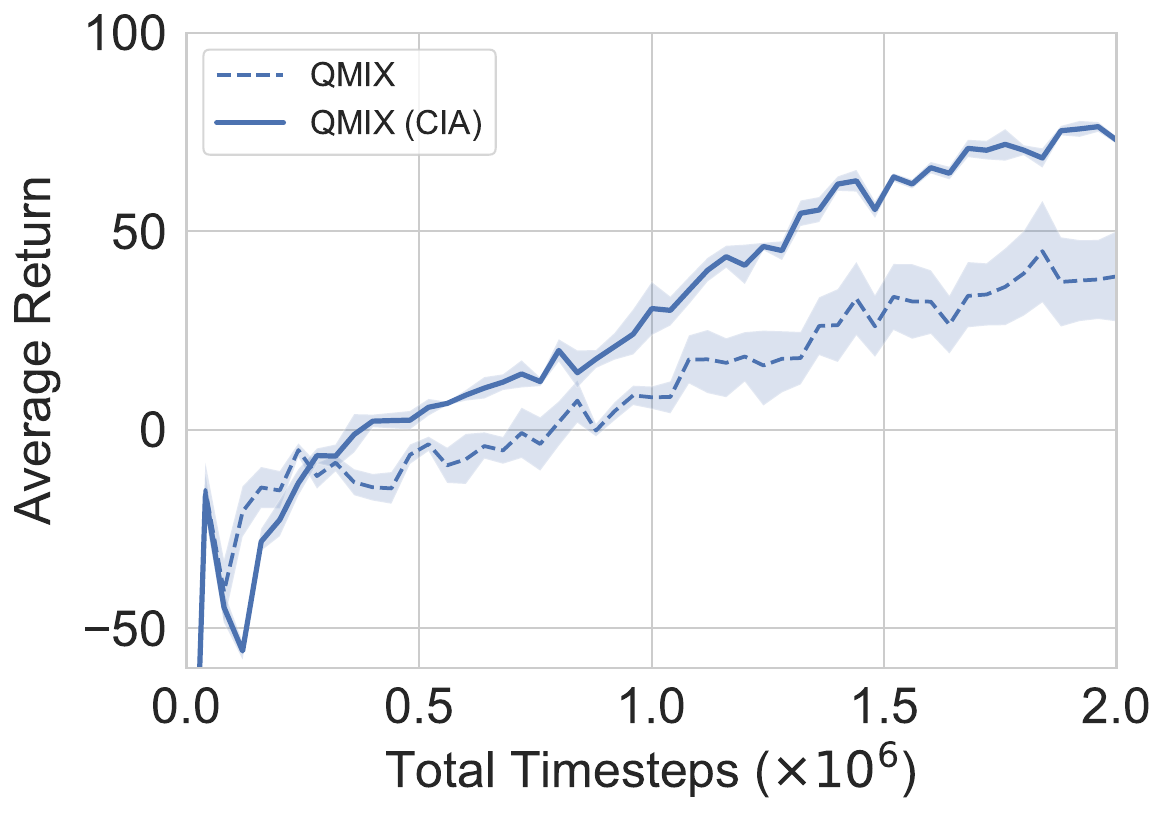}}\;
    \subfloat[Temporal Credits]{\includegraphics[width=0.4\textwidth]{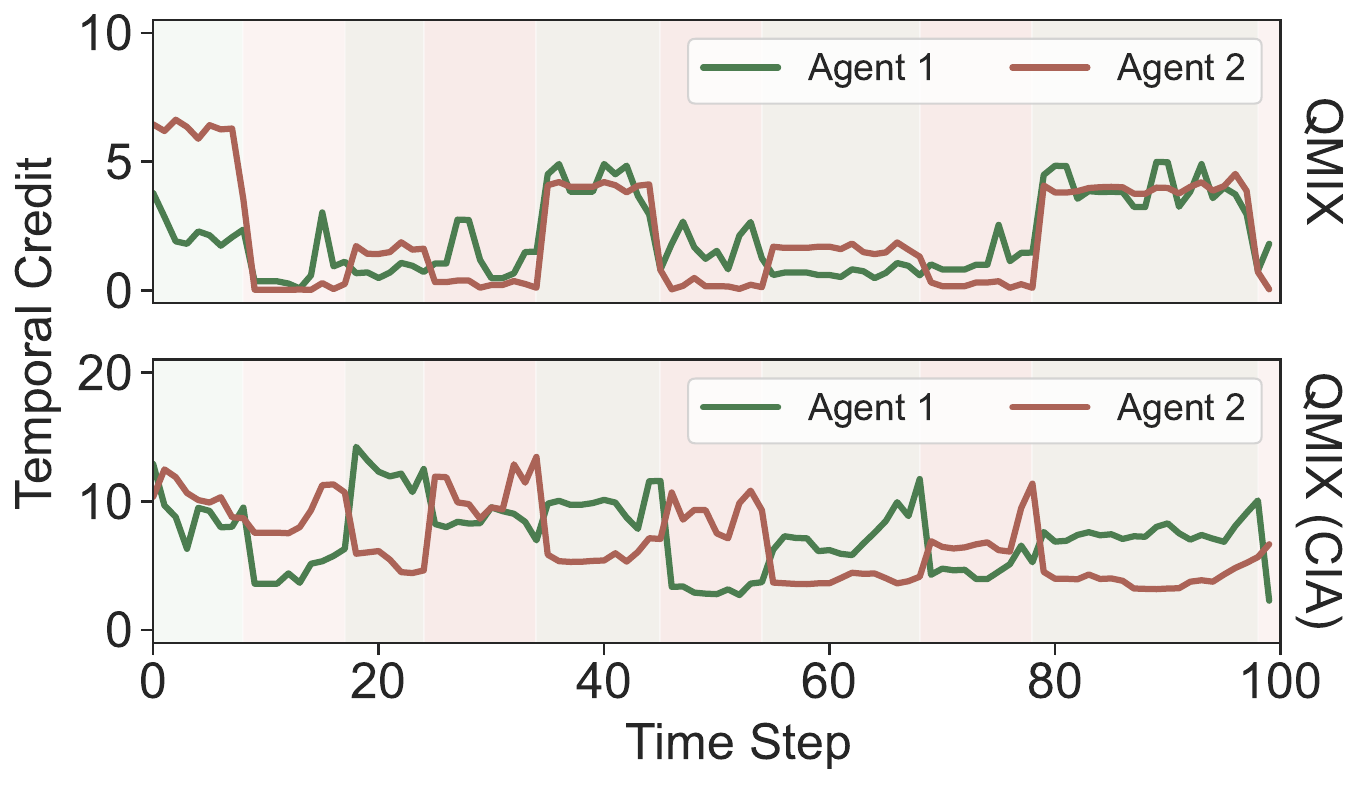}}
    \caption{The performance comparison on the didactic game of Turn. }
    \label{fig:game}
\end{figure*}

\begin{figure*}[!t]
    \centering
    \subfloat{\quad\quad\includegraphics[width=0.90\textwidth]{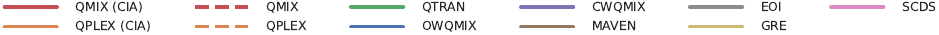}}\\    
    \addtocounter{subfigure}{-1}
    \subfloat[10m\_vs\_11m~(Easy)]{\includegraphics[width=0.33\textwidth]{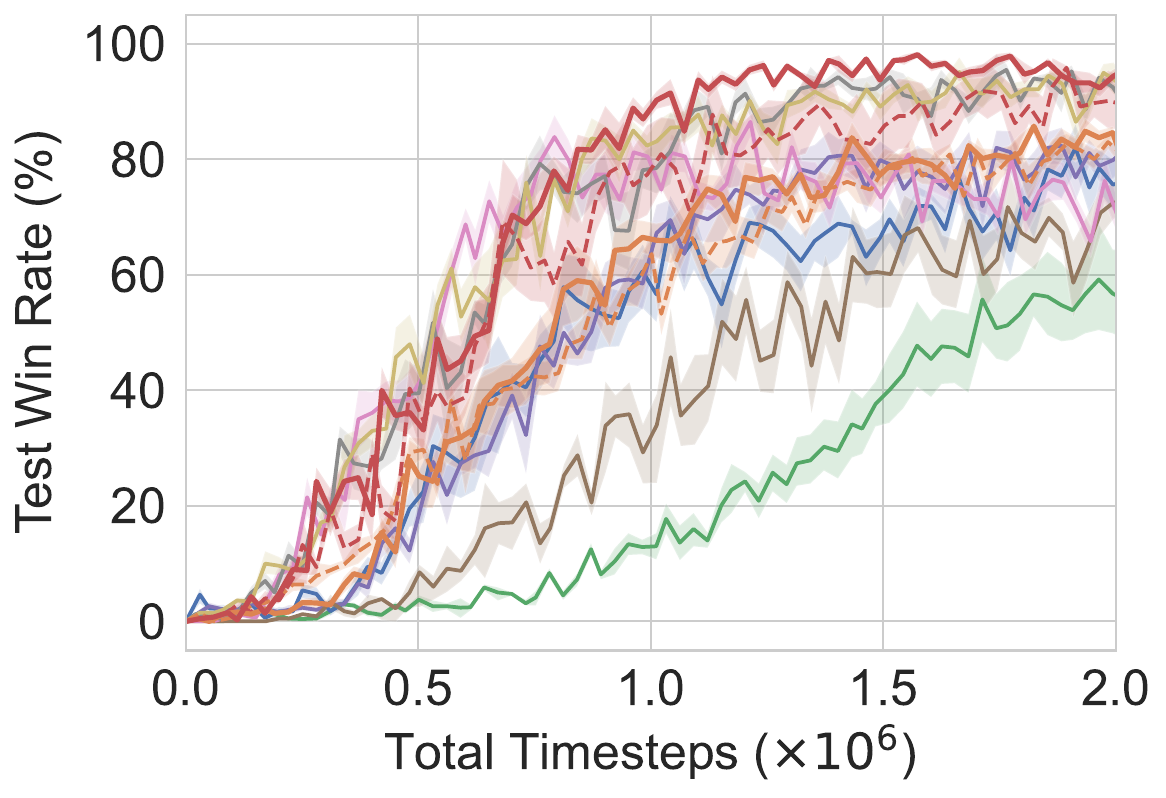}}
    \subfloat[2c\_vs\_64zg~(Hard)]{\includegraphics[width=0.33\textwidth]{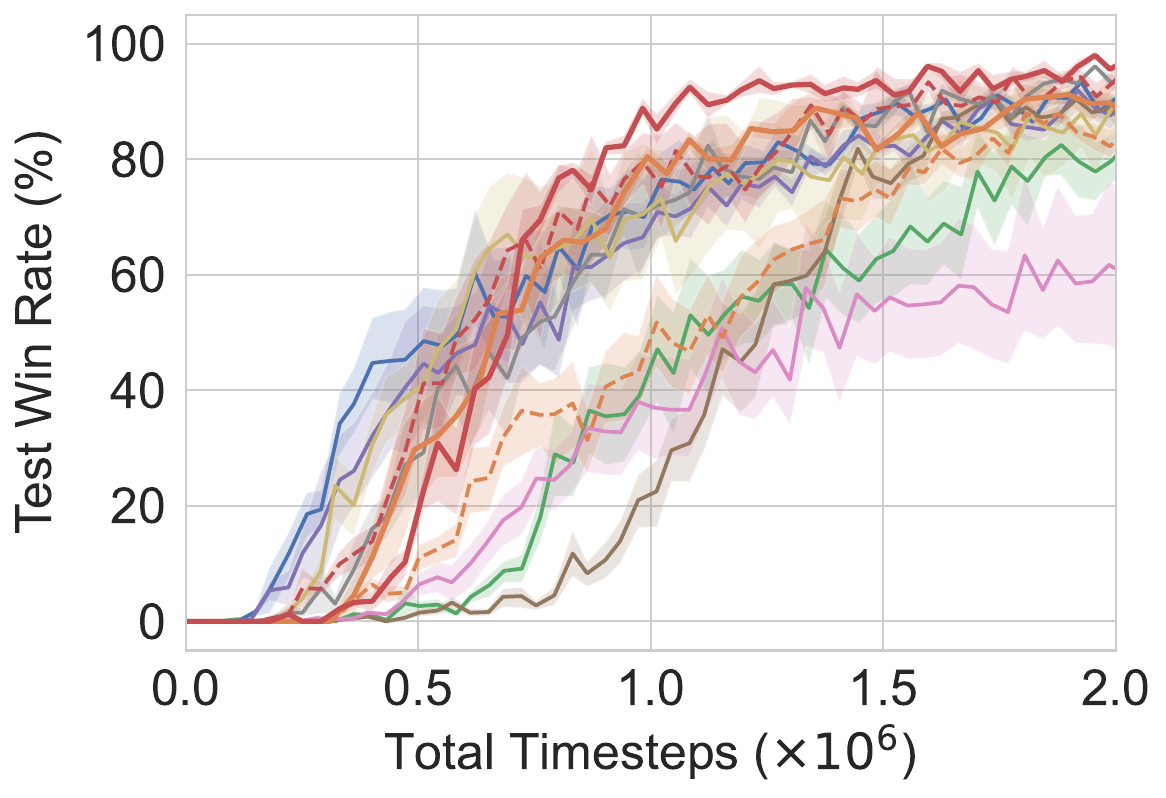}}
    \subfloat[7sz~(Hard)]{\includegraphics[width=0.33\textwidth]{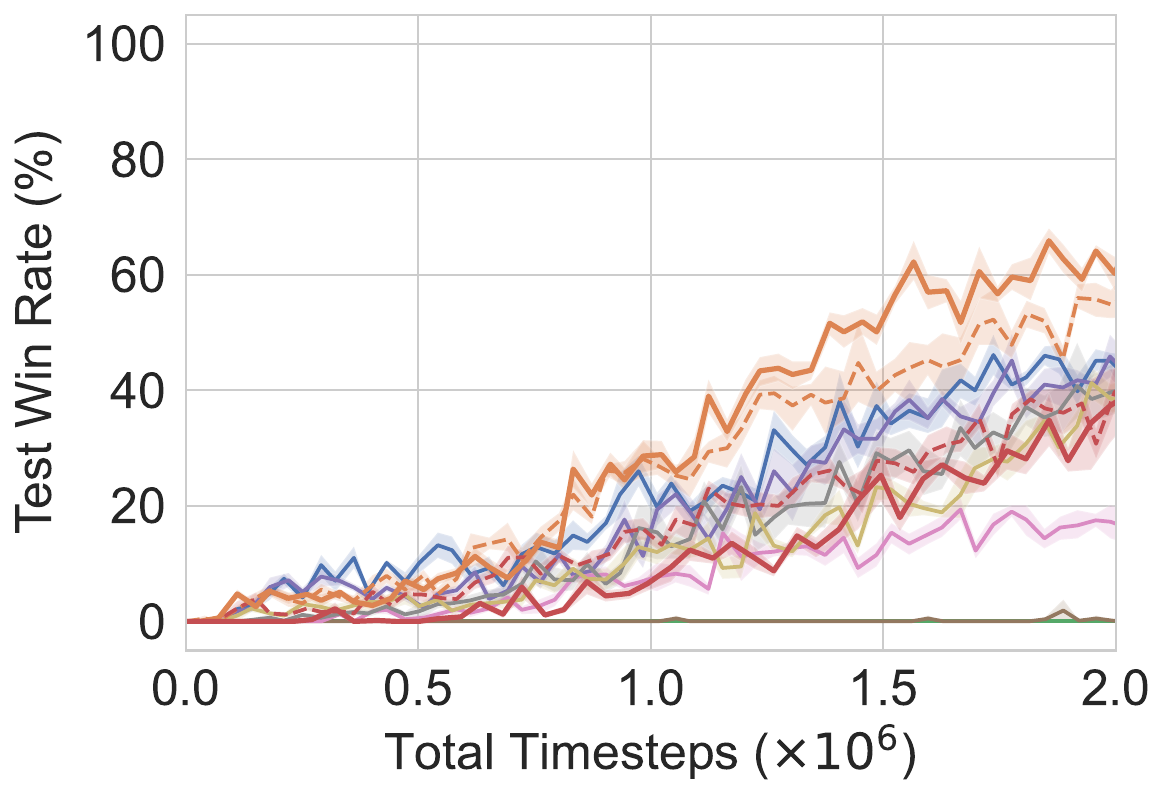}}\\
    \subfloat[6h\_vs\_8z~(Super Hard)]{\includegraphics[width=0.33\textwidth]{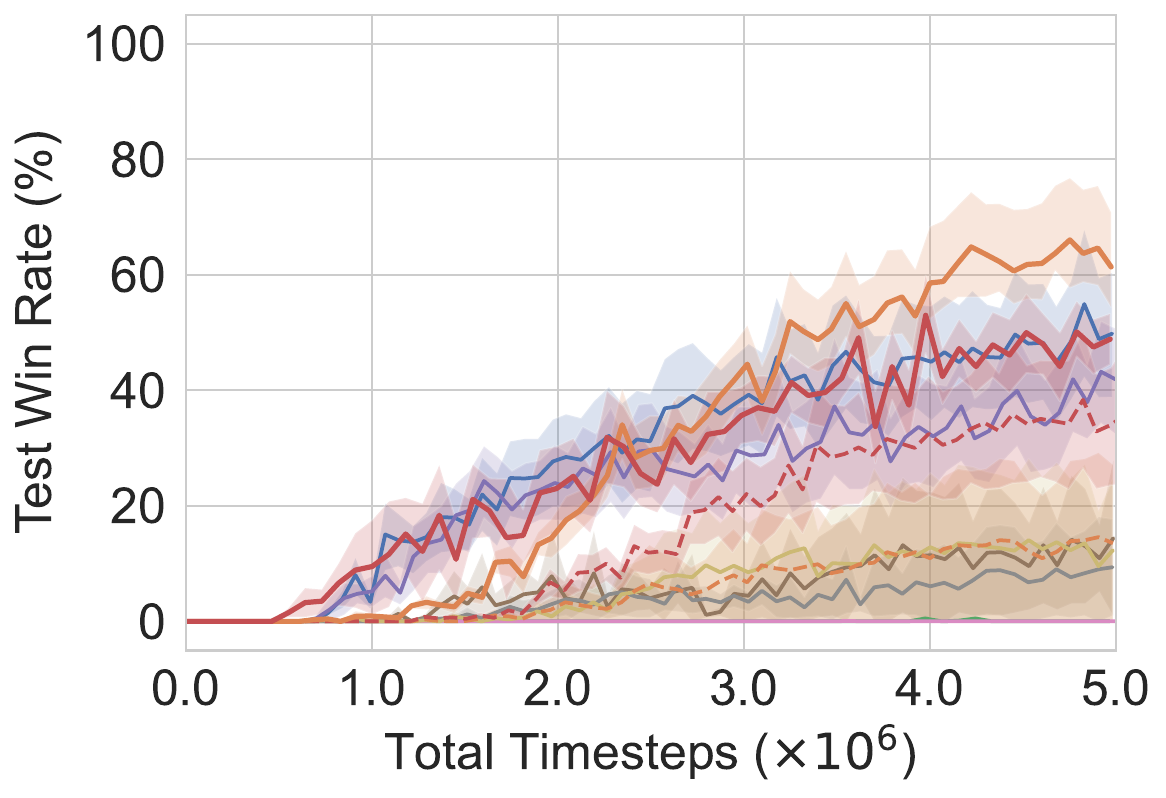}}
    \subfloat[corridor~(Super Hard)]{\includegraphics[width=0.33\textwidth]{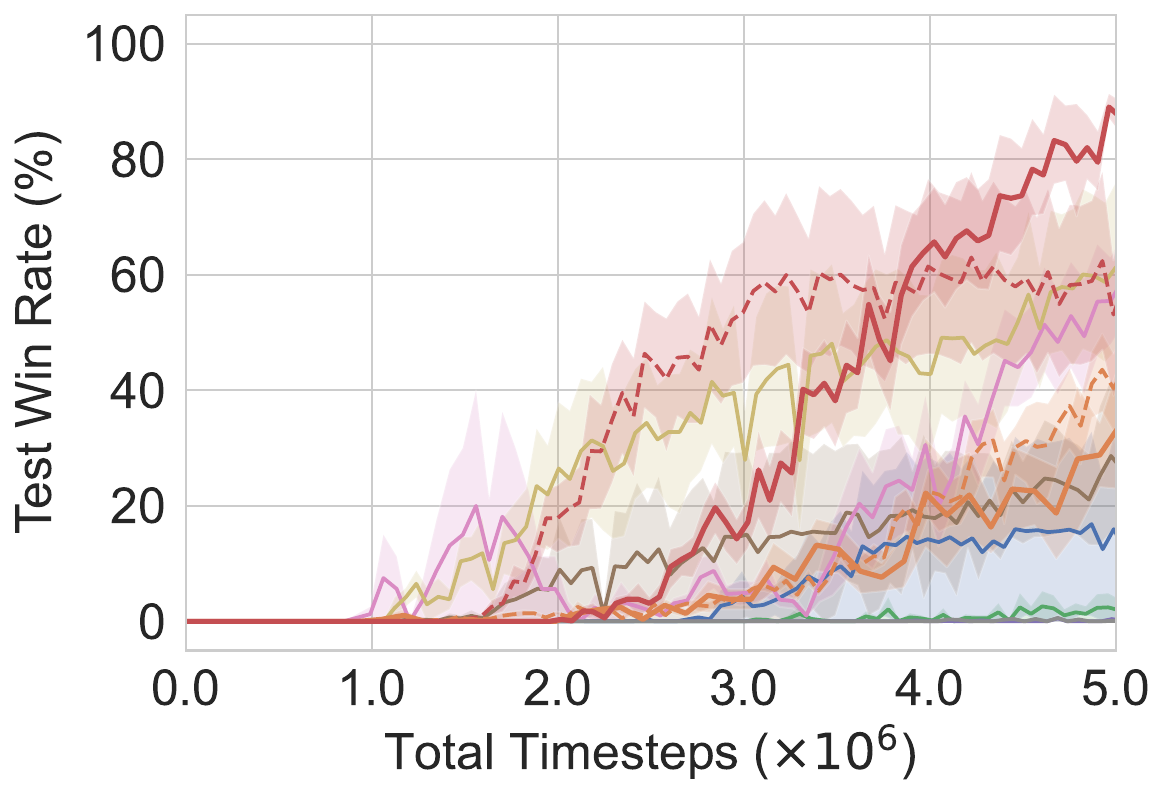}}
    \subfloat[3s5z\_vs\_3s6z~(Super Hard)]{\includegraphics[width=0.33\textwidth]{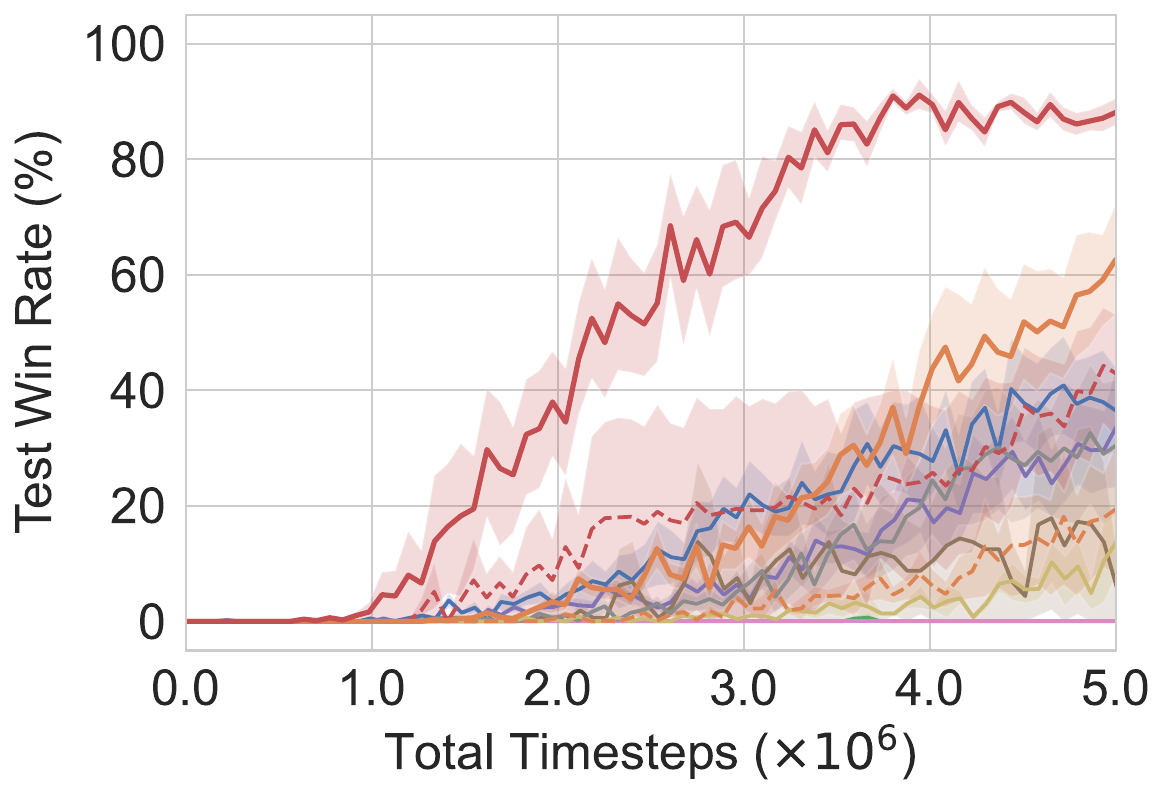}}
    \caption{Learning curves of our proposed CIA variants and baselines on the SMAC scenarios. All experimental results are illustrated with the mean and the standard deviation of the performance over five random seeds for a fair comparison. To make the results in figures clearer for readers, we adopt a 50\% confidence interval to plot the error region.}
    \label{fig:results}
  \end{figure*}

\subsection{Didactic Game} 
We design a didactic game and call it \emph{Turn}. 
Two colored agents with the observation of $3\times3$ are initialized in a $5\times5$ map, and the corresponding colored apples are generated alternately in the map. 
The goal of each agent is to take turns to eat the exclusive apples (agent 1 first), where the one that is not in its round will be trapped by fences and punished when moving. This didactic game requires the VD model to learn a rotating credit assignment strategy. As shown in Figure~\ref{fig:game}, the proposed CIA works quite well on improving QMIX. We further visualize the temporal credits of different methods in a sampled trajectory. Obviously, CIA realizes the distinguishable credit assessment, where the temporal credits of two agents rise alternately. However, the credits of QMIX are sometimes ambiguous, which even pay more attention to the trapped agent in each round. This crisp example verifies the effectiveness of the proposed CIA. Please refer to Appendix B for more details.

\subsection{SMAC Benchmark} 
The StarCraft Multi-Agent Challenge~(SMAC)\footnote{We use SC2.4.10 version instead of the older SC2.4.6.2.69232. Performance is not comparable across versions.}~\cite{SMAC} has become a common-used benchmark for evaluating state-of-the-art MARL methods. SMAC focuses on micromanagement challenges where each of the ally entities is controlled by an individual learning agent, and enemy entities are controlled by a built-in AI. The goal of the agents is to maximize the damage to enemies. Hence, proper tactics such as sneaking attack and drawing fire are required during battles. Learning these diverse interaction behaviors under partial observation is a challenging task.
To validate the effectiveness of our methods, we conduct experiments on 6 SMAC scenarios~\cite{SMAC} which are classified into \textbf{Easy}~(\emph{10m\_vs\_11m}), \textbf{Hard}~(\emph{2c\_vs\_64zg}, \emph{7sz}) and \textbf{Super Hard}~(\emph{6h\_vs\_8z}, \emph{corridor}, \emph{3s5z\_vs\_3s6z}). Only the \emph{10m\_vs\_11m} scenario is homogeneous, where the army is composed of only a single unit type, while the others are heterogeneous.

The experimental results on different scenarios are shown in Figure~\ref{fig:results}. Compared with the state-of-the-art baseline methods, our proposed CIA successfully improves the final performance. In the easy homogeneous scenarios~(\emph{10m\_vs\_11m}), the strength gap between the agents and enemies is small. Thus, several baselines can also achieve promising results without complex tactics, while the benefit of diversity brought by CIA is not obvious. However, in the more difficult heterogeneous scenario (\emph{2c\_vs\_64zg}, \emph{7sz}), our method yields better performance than baselines. The results suggest that CIA can be utilized to provide a more discriminative credit assignment, which helps heterogeneous agents to explore diverse cooperation strategies. Moreover, this exploration may lead to a little drop in learning efficiency, but it is worthwhile as a trade-off for achieving non-trivial performance. To further test the potentiality of the proposed method, we also compare CIA in the super hard heterogeneous scenarios~(\emph{6h\_vs\_8z}, \emph{corridor}, \emph{3s5z\_vs\_3s6z}). In these challenging scenarios, learning distinguishable credits becomes complex due to the different unit types. Furthermore, there is a great disparity in strength between the two teams, and it is impossible to beat the enemies in a reckless way. The proposed CIA provides an impressive improvement in the performance over the baselines, showing its robustness to diversity promoting. Especially in the \emph{3s5z\_vs\_3s6z} scenario, almost all compared baselines cannot learn any effective policy and perform poorly, while our method still maintains the superior performance.

\subsection{Ablation Study} 

To verify the generalization of the proposed method, we apply the CIA module to two VD methods, including QMIX and QPLEX, and present the results in Figure~\ref{fig:results}. In general, the proposed CIA shows promising performance in enhancing different VD methods. Of special interest is the significant improvement in the super hard heterogeneous scenarios~(\emph{3s5z\_vs\_3s6z}), where credit assignment plays an important role. CIA~(QMIX) and CIA~(QPLEX) both obtain gratifying results superior to QMIX and QPLEX, indicating that the CIA module successfully enlarges the credit-level distinguishability. Moreover, it is notable that the improvement of incorporating CIA into different VD methods relies on the representation capability of the original models. The dedicated CIA module focuses on providing a regularization term to guide the optimization of the credit assignment without changing the intrinsic VD mechanism.

\begin{figure}[!t]
    \centering
    \subfloat{\includegraphics[width=0.47\textwidth]{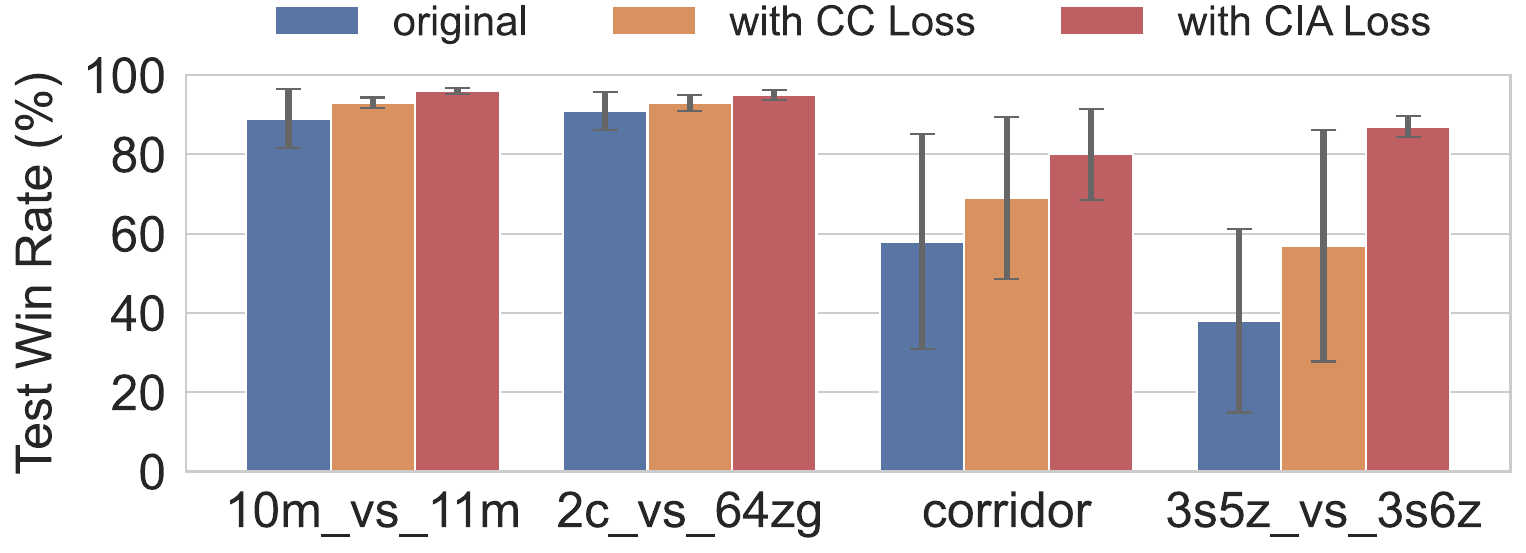}}
    \caption{The performance comparison between the QMIX variants with the CC loss and the CIA loss, respectively.}
    \label{fig:loss}
  \end{figure}

Moreover, to further demonstrate the advantage of CIA in promoting credit-level distinguishability, we also design a comparable variant that uses a credit classification loss~(CC Loss) instead of the proposed contrastive identity-aware learning loss~(CIA Loss). The credit classification loss is based on directly predicting the corresponding identity labels given the temporal credits. Figure~\ref{fig:loss} reports the experimental results of QMIX with different loss items. The dedicated CC loss can also make an improvement to alleviate the ambiguous credit assignment problem. In the simple scenarios~(\emph{10m\_vs\_11m}, \emph{2c\_vs\_64zg}), the CIA loss offers close performance compared with CC loss. However, in the difficult scenarios~(\emph{corridor}, \emph{3s5z\_vs\_3s6z}), the CIA loss outperforms the CC loss by a wide margin, while the CC loss suffers from a large variance. The results show that the CIA loss constraints the temporal credits and the identity representations into a specific credit-identity space without divergence, providing a more stable and robust optimization objective. However, the learned credits of the CC loss may distribute far away from the decision boundary constructed by the identity representations.

\section{Conclusion}

In this paper, we investigate the multi-agent diversity problem of the value decomposition~(VD) paradigm in the MARL setting, observing that existing works on diverse agent networks are limited by the indistinguishable credit assignment of the VD network. Inspired by the observation, we propose a novel contrastive identity-aware learning~(CIA) method to explicitly encourage the VD network to distinguish the assigned credits of multiple agents. Technically, we maximize the mutual information of the temporal credits and the identity representations of different agents with a tractable lower bound. We validate CIA over the StarCraft II micromanagement benchmark, and showcase that it yields results significantly superior to the state-of-the-art techniques in MARL. This simple yet effective CIA method further motivates us to explore the high-quality representation in future work rather than improving the network architecture. Another challenging direction is the theoretical study of the ambiguous credit assignment phenomenon. It is interesting to study the relationship between the IGM constraint and the ambiguous credit assignment problem. The IGM constraint may limit the distinguishable expressiveness of credit assignment in the existing VD architectures.

\section*{Acknowledgments}
This work is funded by the National Key R\&D Program of China (Grant No: 2018AAA0101503) and the Science and technology project of SGCC (State Grid Corporation of China): fundamental theory of human-in-the-loop hybrid-augmented intelligence for power grid dispatch and control.

\bibliography{aaai23}

\newpage

\newpage

\appendix

\section*{Appendix}

In the appendix, we provide additional materials that cannot fit into the main manuscript due to page limit, including pseudocode, experimental settings, and additional results.

\section{Method}

To make the proposed Contrastive Identity-Aware learning~(CIA) method clearer for readers, the pseudocode is provided in Algorithm~\ref{alg:cia}. The algorithm implementation of the CIA module is simple yet effective that can be seamlessly integrated with various VD architectures, which is highlighted with a yellow background\footnote{In QMIX, we can directly derive the gradients instead of the $\operatorname{autograd}$ operation for faster calculation.~\cite{GRE}}. The most importance operation is to transpose the similarity matrix as shown in Line~\#23, avoiding the contrastive identity-aware learning problem degenerating into a credit classification problem that is easily over-fitting.

\begin{algorithm}[!h]
	\caption{CIA Pseudocode~(PyTorch-like)}
	\footnotesize
	\begin{algorithmic}[1]
    \STATE \texttt{\anno{\# unity, mixer: training networks}}
    \STATE \texttt{\anno{\# unity$^-$, mixer$^-$: target networks}}
    \STATE \texttt{\anno{\# batch: trajectories of shape [B,N+1,K]}}
    \STATE \texttt{\anno{\# B: batch size}}
    \STATE \texttt{\anno{\# N: trajectory length}}
    \STATE \texttt{\anno{\# K: agent number}}
    \STATE \texttt{\anno{\# o: observation, s: state, r: reward}}
    \STATE \texttt{\anno{\# $\alpha$: trade-off coefficient of loss terms}}
    \STATE 
    \STATE \texttt{\HiLi identity = nn.Linear(N, K) }
    \STATE \texttt{\HiLi loss = nn.CrossEntropyLoss()}
    \STATE \texttt{\pink{def} train(batch):}
    \STATE \texttt{\quad \anno{\# value decomposition learning}}
    \STATE \texttt{\quad $Q^{k}$ = unity(batch["o"][:,:-1])}
    \STATE \texttt{\quad $Q^{tot}$ = mixer($Q^k$, batch["s"][:,:-1])}
    \STATE \texttt{\quad $\hat{Q}^k$ = unity$^-$(batch["o"][:,1:])}
    \STATE \texttt{\quad $\hat{Q}^{tot}$ = mixer$^-$($\hat{Q}^k$, batch["s"][:,1:])}
    \STATE \texttt{\quad target = batch["r"][:,:-1] + $\gamma$ * $\hat{Q}^{tot}$}
    \STATE \texttt{\quad $\mathcal{L}_{TD}$ = ($Q^{tot}$\,-\,target.\pink{detach()})$^2$.mean()}
    \STATE 
    \STATE \texttt{\HiLi \quad \anno{\# contrastive identity-aware learning}}
    \STATE \texttt{\HiLi \quad $X$ = autograd.grad($Q^{tot}$, $Q^k$)\;\anno{\# [B,N,K]}}
    \STATE \texttt{\HiLi \quad $X$ = $X$.transpose(1, 2)\quad\quad\quad\,\anno{\# [B,K,N]}}
    \STATE \texttt{\HiLi \quad $G$ = identity($X$)\qquad\qquad\qquad\;\;\;\anno{\# [B,K,K]}}
    \STATE \texttt{\HiLi \quad $G$ = $G$.\pink{transpose(1, 2)}\qquad\quad\;\,\anno{\# [B,K,K]}}
    \STATE \texttt{\HiLi \quad $G$ = $G$.reshape(-1, K)\qquad\qquad\anno{\# [B*K,K]}}
    \STATE \texttt{\HiLi \quad label = torch.arange(0, K)\;\;\;\anno{\# [K]}}
    \STATE \texttt{\HiLi \quad label = label.repeat(B)\qquad\;\;\,\anno{\# [B*K]}}
    \STATE \texttt{\HiLi \quad $\mathcal{L}_{CL}$ = loss($G$, label)}
    \STATE \texttt{\HiLi \quad $\mathcal{L}_{all}$ = $\mathcal{L}_{TD}$ + $\alpha$ * $\mathcal{L}_{CL}$}
    \STATE 
    \STATE \texttt{\quad \anno{\# update networks}}
    \STATE \texttt{\quad optimiser.zero\_grad()}
    \STATE \texttt{\quad $\mathcal{L}_{all}$.\pink{backward()}}
    \STATE \texttt{\quad optimiser.step()}
    \STATE \texttt{\quad \pink{if} target update interval reached:}
    \STATE \texttt{\quad \quad unity$^-$, mixer$^-$ = unity, mixer}
	\end{algorithmic}
	\label{alg:cia}
\end{algorithm}

\section{Experiments}

To demonstrate the effectiveness of the proposed CIA method, we provide case studies to evaluate the credit indistinguishability of existing VD methods. Then we conduct experiments on the didactic game of \emph{Turn} and the StarCraft II micromanagement challenge~\cite{SMAC}.

\subsection{Experimental Settings}

\paragraph*{Baselines.} Our methods are compared with various state-of-the-art methods, including (i) Value decomposition methods: QMIX~\cite{QMIX}, QPLEX~\cite{QPLEX}, QTRAN~\cite{QTRAN}, OWQMIX and CWQMIX~\cite{WQMIX}. (ii) Diversity-based methods: MAVEN~\cite{MAVEN}, EOI~\cite{EOI}, GRE~\cite{GRE}, SCDS that is a variant of CDS~\cite{CDS} with complete shared agent network to ensure comparability. 

\paragraph*{Architectures.} Our proposed CIA implementation uses QMIX~\cite{QMIX} and QPLEX~\cite{QPLEX} as the integrated backbones to evaluate its performance, namely QMIX~(CIA) and QPLEX~(CIA). These two methods are chosen for their robust performance in different multi-agent tasks, while CIA can also be readily applied to other methods. The detailed network architectures are referred in their source codes. We only introduce an additional linear layer for learnable identity representations with the shape of $[N,K]$, where $N$ is the maximum length of trajectories and $K$ is the number of agents both determined by the specific scenarios.

\paragraph*{Hyperparameters.} We adopt the Python MARL framework (PyMARL)~\cite{SMAC} to implement our method and all baselines. The detailed hyperparameters in the didactic game and the SMAC benchmark are given as follows, where the common hyperparameters across different methods are consistent for a fair comparison. We modify several hyperparameters in some difficult scenarios and also detail them below. 

All experimental results are illustrated with the mean and the standard deviation of the performance over five random seeds for a fair comparison. To make the results in figures clearer for readers, we adopt a 50\% confidence interval to plot the error region. Batches of 32 episodes are sampled from the replay buffer with the size of 5K every training iteration. The target update interval is set to 200, and the discount factor is set to 0.99. We use the RMSprop optimizer with a learning rate of $5 \times 10^{-4}$, a smoothing constant of 0.99, and with no momentum or weight decay. For exploration, $\epsilon$-greedy is used with $\epsilon$ annealed linearly from 1.0 to 0.05 over 50K timesteps and kept constant for the rest of the training. In several super hard SMAC scenarios~(\emph{6h\_vs\_8z}, \emph{corridor}, \emph{3s5z\_vs\_3s6z}) that require more exploration, the test win rates often remain 0\% with the default hyperparameters. In this way, we extend the epsilon anneal time to 500K, and two of them~(\emph{6h\_vs\_8z}, \emph{corridor}) optimized with the Adam optimizer for all the compared methods. 

For the CIA module, the additional hyperparameter is the trade-off coefficient $\alpha$ of the CIA loss. We search $\alpha$ over $\{0.005, 0.01, 0.02, 0.04, 0.05, 0.1, 0.5, 1.0\}$. In the didactic game of \emph{Turn}, the coefficient $\alpha$ of QMIX~(CIA) is set to 0.02. 
In the SMAC benchmark, the coefficient $\alpha$ of QMIX~(CIA) is set to 0.02, 0.02, 0.005, 0.005, 0.02, and 0.1 for the \emph{10m\_vs\_11m}, \emph{2c\_vs\_64zg}, \emph{7sz}, \emph{6h\_vs\_8z}, \emph{corridor}, and \emph{3s5z\_vs\_3s6z} scenarios, respectively, while the coefficient $\alpha$ of QPLEX~(CIA) is set to 0.1, 0.02, 0.02, 0.1, 0.02, and 0.04, respectively. In the ablation study, the coefficient $\alpha$ of QMIX~(CC) is consistent with the above QMIX~(CIA). Experiments are carried out on NVIDIA Quadro P5000 GPU.

\subsection{Credit Indistinguishability Analysis}

To investigate the ambiguous credit assignment problem in VD, we propose to use KL-divergence distance to measure the similarity of credit distributions of different methods. For each compared method, we first use the corresponding trained network to sample trajectories. All the trajectories sampled by different methods are mixed together and used for following calculations to ensure comparability. Then we calculate the average KL-divergence distance of credit distributions for each pair of the compared methods in all sampled trajectories. Specially, given a trained network, the credit of the agent $a_k$ at each time step $t$ of the trajectories is formulated as: 
\begin{align}
    x^k_t = \frac{\partial Q^{tot}_t}{\partial Q^k_t} \in \mathbb{R}.
\end{align}
Thus, the credit distribution $\boldsymbol{d}_t$ is defined as the normalized gradient-based credits over all $K$ agents: 
\begin{align}
    \boldsymbol{d}_t = \operatorname{softmax}\left( \left[x^1_t,x^2_t,\cdots,x^K_t\right] \right).
\end{align}
The average KL-divergence distance $\Lambda$ of credit distributions between the trained network $i$ and $j$ is calculated as:
\begin{align}
    \Lambda (i, j) = \mathbb{E}_{\mathcal{B}}\left[ \operatorname{KL}(d_t^i || d_t^j) \right],
\end{align}
where $\operatorname{KL}(\cdot)$ is the KL-divergence and $\mathcal{B}$ denotes the set of all sampled trajectories. Here each compared method sample $10$ trajectories. A total of $30$ trajectories are obtained from $3$ compared methods, including QMIX, QMIX~(RS) and QMIX~(CIA). Please refer to the main manuscript for the results of the \emph{2c\_vs\_64zg} and \emph{3s5z\_vs\_3s6z} scenarios.

\subsection{Didactic Game}

\begin{figure}[!h]
    \centering
    \includegraphics[width=0.40\textwidth]{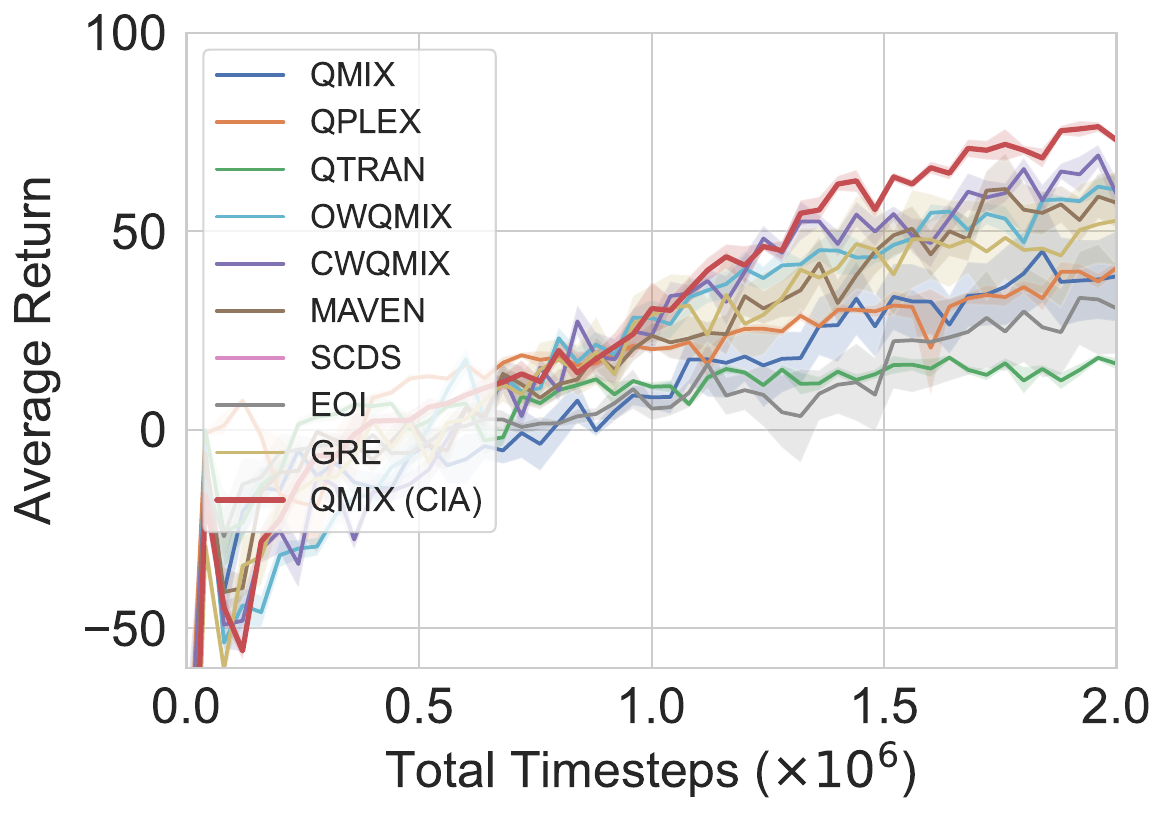}
    \caption{Learning curves of our method and baselines on the didactic game of Turn.}
    \label{fig:turn}
  \end{figure}

To verify the effectiveness of the proposed CIA method, we design a didactic game and call it \emph{Turn}. 
Two colored agents with the observation of $3\times3$ are initialized in a $5\times5$ map, and the corresponding colored apples are generated alternately in the map. Only when the current apple is eaten, the next apple will be generated. The goal of each agent is to take turns to eat the exclusive apples (agent 1 first), where the one that is not in its round will be trapped by fences and punished when moving. At each time step, agents can move in four cardinal directions, stay, or eat. Thus, the action space for each agent consists of 6 discrete actions. Agents will receive a global reward of 10 if the free agent eats the corresponding apple. On the other hand, agents will be punished by -1 if the trapped agent moves or the free agent move to hit the borders of the map. The total length of each trajectory is 100. This didactic game requires the VD model to learn a rotating credit assignment strategy. Please refer to the main manuscript for the main results and analysis. Here we provide the additional compared results of all baselines on the didactic game of Turn, as shown in Figure~\ref{fig:turn}. The proposed CIA works quite well on improving the final performance compared with the state-of-the-art baselines.

\subsection{SMAC Benchmark}

The StarCraft Multi-Agent Challenge~(SMAC)\footnote{We use SC2.4.10 version instead of the older SC2.4.6.2.69232. Performance is not comparable across versions.}~\cite{SMAC} has become a common-used benchmark for evaluating state-of-the-art MARL methods. SMAC focuses on micromanagement challenges where each of the ally entities is controlled by an individual learning agent, and enemy entities are controlled by a built-in AI. At each time step, agents can move in four cardinal directions, stop, take no-operation, or choose an enemy to attack. Thus, if there are $N_e$ enemies in the scenario, the action space for each ally unit consists of $N_e + 6$ discrete actions. The goal of the agents is to maximize the damage to enemies. Hence, proper tactics such as sneaking attack and drawing fire are required during battles. Learning these diverse interaction behaviors under partial observation is a challenging task.

To validate the effectiveness of our methods, we conduct experiments on 6 SMAC scenarios~\cite{SMAC} which are classified into \textbf{Easy}~(\emph{10m\_vs\_11m}), \textbf{Hard}~(\emph{2c\_vs\_64zg}, \emph{7sz}) and \textbf{Super Hard}~(\emph{6h\_vs\_8z}, \emph{corridor}, \emph{3s5z\_vs\_3s6z}). Please refer to the main manuscript for the learning curves of compared methods. Here we provide another quantitative metric to compare different methods. We calculate the average test win rate over the last 500K training steps for each model due to the unstable fluctuation of performance. The additional quantitative results are reported in Table~\ref{tab:result}. Compared with the state-of-the-art baseline methods, our proposed CIA successfully improves the final performance. The results suggest that CIA can be utilized to provide a more discriminative credit assignment, which helps agents to explore diverse cooperation strategies.

\section{Visualization}

To further demonstrate the effectiveness of the proposed CIA, we provide several visualization examples of the diverse strategies that emerge in different SMAC scenarios. Green and red shadows indicate the agent and enemy formations, respectively. Green and red arrows indicate the moving direction of the agent and enemy formations, respectively. White arrows indicate the attack action from the agent units to the enemy units.

\begin{figure*}[!h]
    \vspace{-0.1cm}
    \centering
    \includegraphics[width=1.0\textwidth]{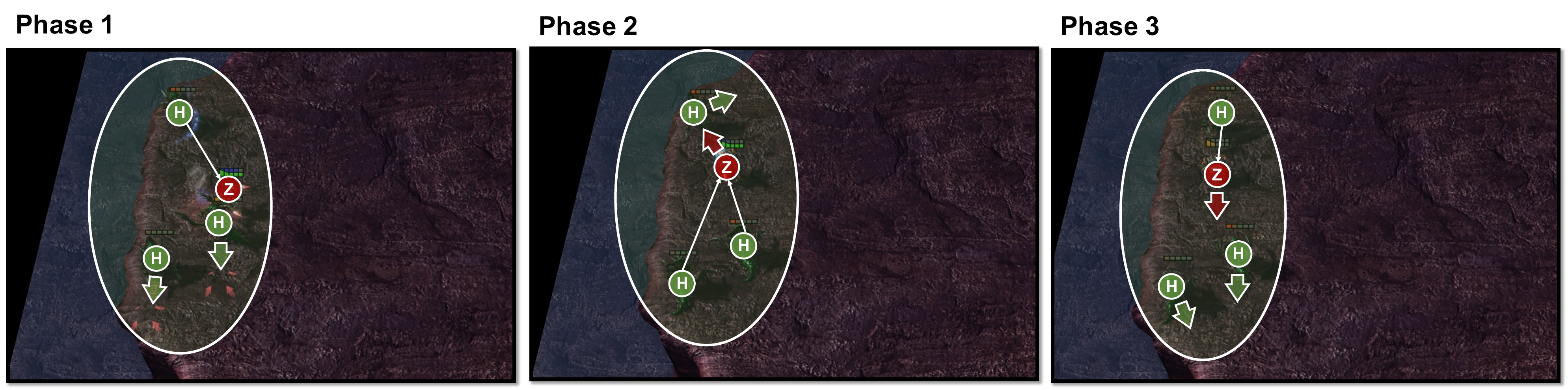}
    \caption{A visualization example of the diverse strategies that emerge in the \emph{6h\_vs\_8z} scenario. }
    \label{fig:v1}
  \end{figure*}

\begin{figure*}[!h]
    \vspace{-0.2cm}
    \centering
    \includegraphics[width=1.0\textwidth]{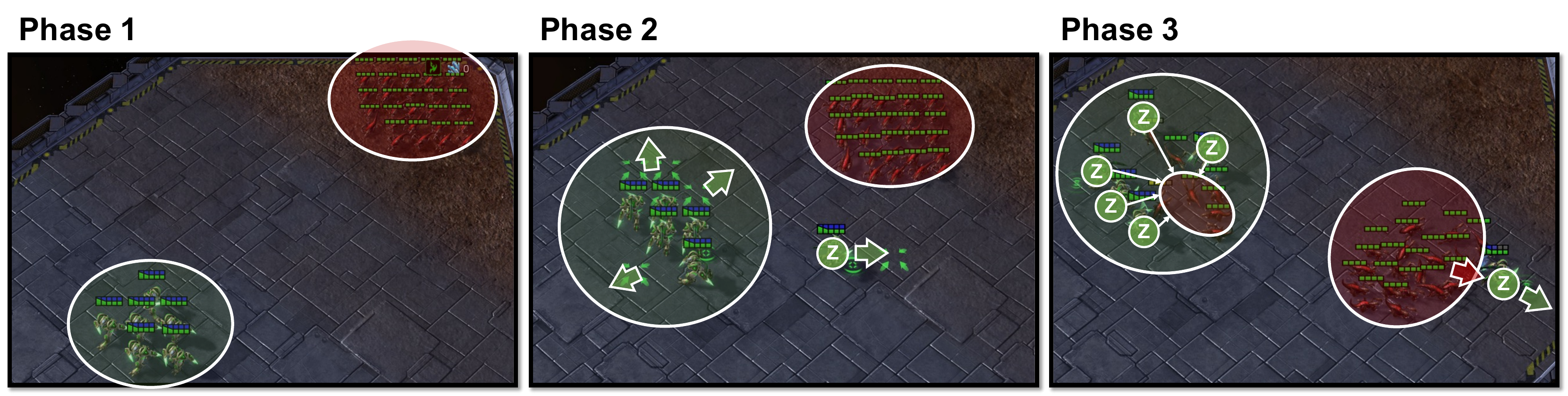}
    \caption{A visualization example of the diverse strategies that emerge in the \emph{corridor} scenario. }
    \label{fig:v2}
  \end{figure*}

\begin{itemize}
    \item As shown in Figure~\ref{fig:v1}, agents master an interesting \emph{Alternate Attack} strategy in the super hard heterogeneous \emph{6h\_vs\_8z} scenario. Three agents first keep moving to ensure a safe distance to the enemy. Then they perform attacks alternately to attract the enemy. Thus, the enemy sometimes moves up and sometimes moves down, wasting a lot of time on moving. The agents successfully fool and kill the enemy via this \emph{Alternate Attack} strategy, without getting too much hurt.
    \item Furthermore, agents master the \emph{Kite} strategy and the \emph{Encircle} strategy in the super hard heterogeneous \emph{corridor} scenario as shown in Figure~\ref{fig:v2}. At the beginning of the battle, one agent first leave the team separately to attract the attention of most enemies. This agent successfully kites the enemies and draws the fire to cover the other agents. Then the other agents move in different directions to form encirclement around the remaining enemies. 
    \item In the super hard heterogeneous \emph{3s5z\_vs\_3s6z} scenario, agents also master the \emph{Kite} strategy and the \emph{Encircle} strategy. Their learned strategies are largely dependent on the unit types of agents. One Stalker learns to kite the enemies due to its high defense and a large shooting range. Three Zealots tend to move together and focus fire on the enemies, which bypasses the disadvantages of low defense and low attack damage. Moreover, the other Stalkers and Zealots learn to cooperate with each other to attack the enemies in the encirclement. 
\end{itemize}

\begin{table*}[!t]
    \vspace{-0.3cm}
    \centering
    \caption{The test win rate of compared methods on the SMAC scenarios. $\pm$ corresponds to one standard deviation of the average evaluation over 5 trials. \textbf{Bold} indicates the best performance in each SMAC scenario.}
    \label{tab:result}
    \resizebox{\textwidth}{!}{%
    \begin{tabular}{@{}lcccccc@{}}
    \toprule
     \multicolumn{1}{c}{\textbf{Method}} & \textbf{10m\_vs\_11m}  & \textbf{2c\_vs\_64zg}    & \textbf{7sz}          & \textbf{6h\_vs\_8z}      & \textbf{corridor}    & \textbf{3s5z\_vs\_3s6z}  \\ \midrule
    \textbf{QMIX}~\cite{QMIX}  & 0.89 $\pm$ 0.11  & 0.91 $\pm$ 0.07  & 0.33 $\pm$ 0.05  & 0.36 $\pm$ 0.36  & 0.58 $\pm$ 0.40  & 0.38 $\pm$ 0.34  \\ \specialrule{0em}{1pt}{1pt}
    \textbf{QPLEX}~\cite{QPLEX}  & 0.80 $\pm$ 0.04  & 0.83 $\pm$ 0.07  & 0.49 $\pm$ 0.08  & 0.13 $\pm$ 0.29  & 0.38 $\pm$ 0.22  & 0.16 $\pm$ 0.21  \\ \specialrule{0em}{1pt}{1pt}
    \textbf{QTRAN}~\cite{QTRAN}  & 0.51 $\pm$ 0.20  & 0.75 $\pm$ 0.18  & 0.00 $\pm$ 0.00  & 0.00 $\pm$ 0.00  & 0.02 $\pm$ 0.03  & 0.00 $\pm$ 0.00  \\ \specialrule{0em}{1pt}{1pt}
    \textbf{OWQMIX}~\cite{WQMIX}  & 0.73 $\pm$ 0.10  & 0.89 $\pm$ 0.02  & 0.40 $\pm$ 0.06  & 0.49 $\pm$ 0.32  & 0.15 $\pm$ 0.33  & 0.39 $\pm$ 0.22  \\ \specialrule{0em}{1pt}{1pt}
    \textbf{CWQMIX}~\cite{WQMIX}  & 0.79 $\pm$ 0.10  & 0.87 $\pm$ 0.03  & 0.39 $\pm$ 0.05  & 0.39 $\pm$ 0.30  & 0.00 $\pm$ 0.00  & 0.29 $\pm$ 0.22  \\ \specialrule{0em}{1pt}{1pt}
    \textbf{MAVEN}~\cite{MAVEN}  & 0.65 $\pm$ 0.16  & 0.86 $\pm$ 0.03  & 0.00 $\pm$ 0.00  & 0.12 $\pm$ 0.23  & 0.24 $\pm$ 0.35  & 0.13 $\pm$ 0.26  \\ \specialrule{0em}{1pt}{1pt}
    \textbf{EOI}~\cite{EOI}  & 0.92 $\pm$ 0.03  & 0.92 $\pm$ 0.02  & 0.34 $\pm$ 0.11  & 0.08 $\pm$ 0.17  & 0.00 $\pm$ 0.00  & 0.30 $\pm$ 0.34  \\ \specialrule{0em}{1pt}{1pt}
    \textbf{GRE}~\cite{GRE}  & 0.91 $\pm$ 0.04  & 0.85 $\pm$ 0.18  & 0.28 $\pm$ 0.06  & 0.12 $\pm$ 0.28  & 0.56 $\pm$ 0.40  & 0.09 $\pm$ 0.11  \\ \specialrule{0em}{1pt}{1pt}
    \textbf{SCDS}~\cite{CDS}  & 0.73 $\pm$ 0.03  & 0.58 $\pm$ 0.30  & 0.15 $\pm$ 0.04  & 0.00 $\pm$ 0.00  & 0.52 $\pm$ 0.28  & 0.00 $\pm$ 0.00  \\ \specialrule{0em}{1pt}{1pt} \cmidrule(l){1-7} 
    \textbf{QMIX (CIA)}  & \textbf{0.95 $\pm$ 0.01}  & \textbf{0.95 $\pm$ 0.02}  & \text{0.29 $\pm$ 0.10}  & \text{0.49 $\pm$ 0.16}  & \textbf{0.80 $\pm$ 0.17}  & \textbf{0.87 $\pm$ 0.04}  \\ \specialrule{0em}{1pt}{1pt}
    \textbf{QPLEX (CIA)}  & \text{0.82 $\pm$ 0.04}  & \text{0.88 $\pm$ 0.04}  & \textbf{0.58 $\pm$ 0.05}  & \textbf{0.63 $\pm$ 0.28}  & \text{0.27 $\pm$ 0.13}  & \text{0.55 $\pm$ 0.31}  \\ \specialrule{0em}{1pt}{1pt} \bottomrule
    \end{tabular}%
    }
    \end{table*}

  \begin{figure*}[!h]
    \vspace{-0.2cm}
    \centering
    \includegraphics[width=1.0\textwidth]{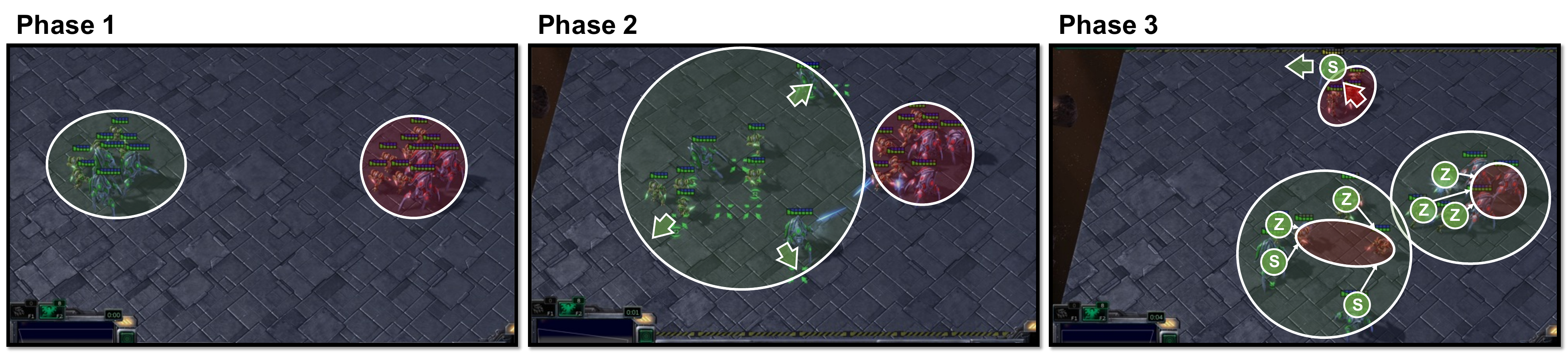}
    \caption{A visualization example of the diverse strategies that emerge in the \emph{3s5z\_vs\_3s6z} scenario. }
    \label{fig:v3}
  \end{figure*}

\end{document}